\documentclass[10pt,twocolumn,letterpaper]{article}

\usepackage{cvpr}
\usepackage{times}
\usepackage{epsfig}
\usepackage{graphicx}
\usepackage{amsmath}
\usepackage{amssymb}

\usepackage{algorithm}
\usepackage{multirow}

\usepackage[breaklinks=true,bookmarks=false]{hyperref}

\cvprfinalcopy 

\def\cvprPaperID{****} 

\begin{document}

\title{Trilaminar Multiway Reconstruction Tree for Efficient Large Scale \\Structure from Motion}

\author{Kun Sun, Wenbing Tao\\
National Key Laboratory of Science and Technology on Multi-spectral Information Processing\\
School of Automation, Huazhong University of Science and Technology\\
Wuhan, 430074, China\\
{\tt\small sunkun@hust.edu.cn, wenbingtao@hust.edu.cn}
}

\maketitle

\begin{abstract}
   Accuracy and efficiency are two key problems in large scale incremental Structure from Motion (SfM). In this paper, we propose a unified framework to divide the image set into clusters suitable for reconstruction as well as find multiple reliable and stable starting points. Image partitioning performs in two steps. First, some small image groups are selected at places with high image density, and then all the images are clustered according to their optimal reconstruction paths to these image groups. This promises that the scene is always reconstructed from dense places to sparse areas, which can reduce error accumulation when images have weak overlap. To enable faster speed, images outside the selected group in each cluster are further divided to achieve a greater degree of parallelism. Experiments show that our method achieves significant speedup, higher accuracy and better completeness.
\end{abstract}

\section{Introduction}
Reconstructing 3D models using online images is a challenging task due to the large scale image set, unknown scene overlap and uncalibrated camera parameters. Such kind of images are usually reconstructed with the Structure from Motion(SfM) method. The most common SfM method operates in an incremental fashion, which consists of three steps: 1) Image matching. In this step, features are extracted and matched between images. Afterwards geometry verification is performed to remove bad matches. 2) Initial model reconstruction. Two starting images having the largest number of matches, subject to the condition that they can not be well modeled by a single homography are selected to build the initial model. 3) Incrementally adding new images. The pose of a new camera is estimated by solving the Perspective-n-Point (PnP) \cite{Lepetit09} problem and then refined with the bundle adjustment algorithm \cite{Triggs00}.

\begin{figure}[t]
	\begin{center}
		\includegraphics[width=0.75\linewidth]{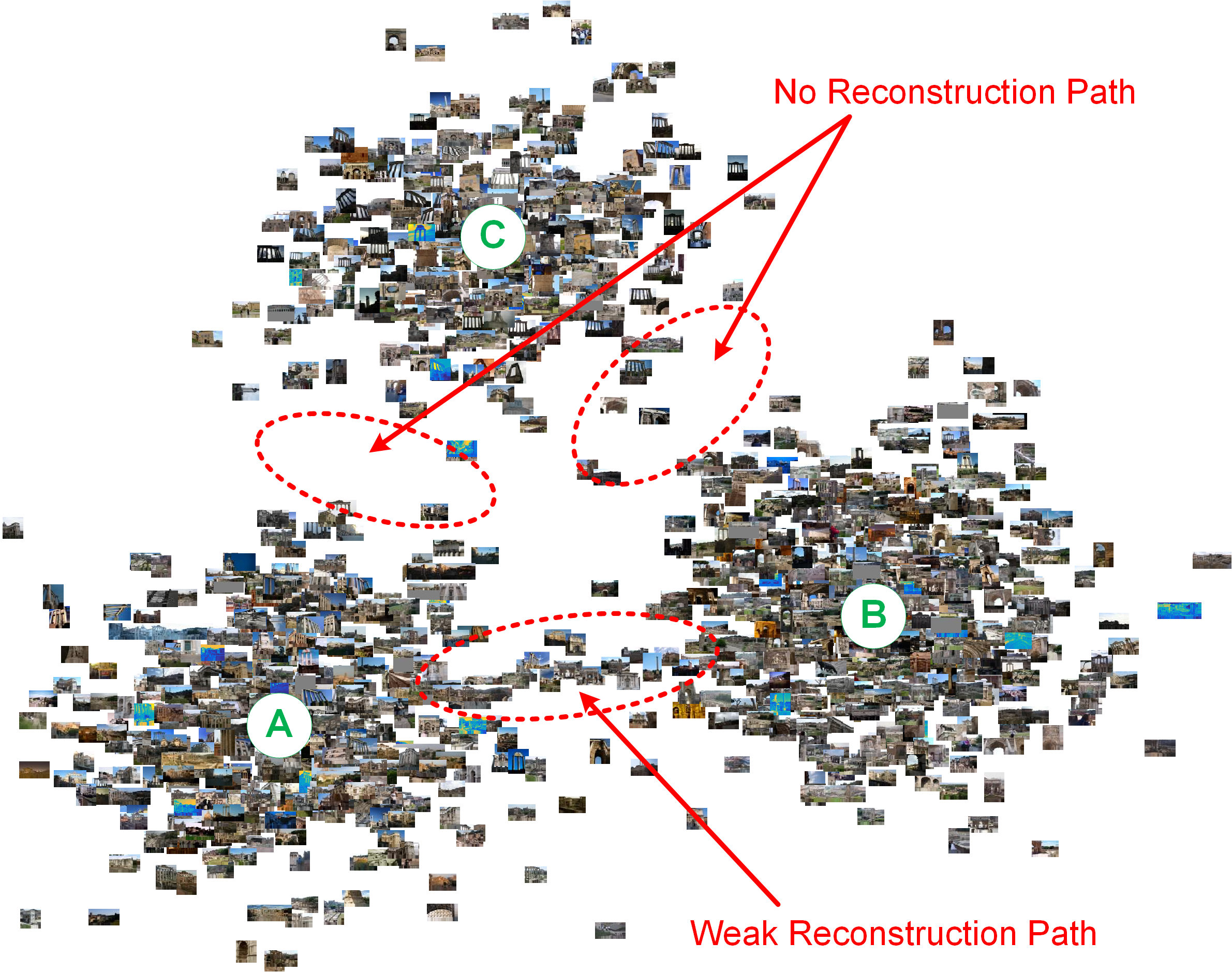}
	\end{center}
	\caption{A common situation in large scale SfM with Internet images. Images are dense at A, B and C but sparse at other places. A and B are connected by a weak reconstruction path. But there is no sufficient overlap between AC and BC.}
	\label{fig:problem}
\end{figure}

While the above pipeline has broad applications on small and medium problems, it is awkward when dealing with large image sets. Fig. \ref{fig:problem} shows a common situation in large scale SfM with Internet photos. Images are dense at places A, B and C but sparse at other places. A and B are connected by a reconstruction path, which is composed of a series of overlapping images between them. But such a reconstruction path is missing between AC and BC. The performance of traditional methods is largely affected by the uncertainty of starting point selection. For example, if the starting point is selected in A, both A and B could be reconstructed. However, there might be large accumulation error since the overlap between them is weak. If the starting point is found in C, neither A nor B could be reconstructed. Some existing systems \cite{Changchang_visualsfm,schoenberger2016sfm} tackle this problem by restarting a new SfM procedure from the remaining images. However, good models might be reconstructed after many failures, which wastes a lot of time. To overcome these shortcomings, some methods divide the original image set and reconstruct each part independently. The images can be clustered according to their location manually. But this is difficult for Internet images because they are totally unordered and not all of them are tagged with geo-information. Or one can find connected components in the image matching graph \cite{Heinly15}. In this case, A and B are in the same component and the accumulation error will not be eliminated. Normalized Cuts is used to partition the image set automatically in \cite{Li2008iconic}. But the number of clusters must be specified by the user, which is hard to know in advance. Some other methods extract iconic images \cite{Li2008iconic,Frahm10,Heinly15,schoenberger2016sfm} or skeletal graphs \cite{Snavely082,Agarwal09} from the original image set to speed up the reconstruction. However, error accumulation and reconstruction interruption may still happen if the starting point is not well selected after image sampling.

\begin{figure}[t]
	\begin{center}
		\includegraphics[width=0.75\linewidth]{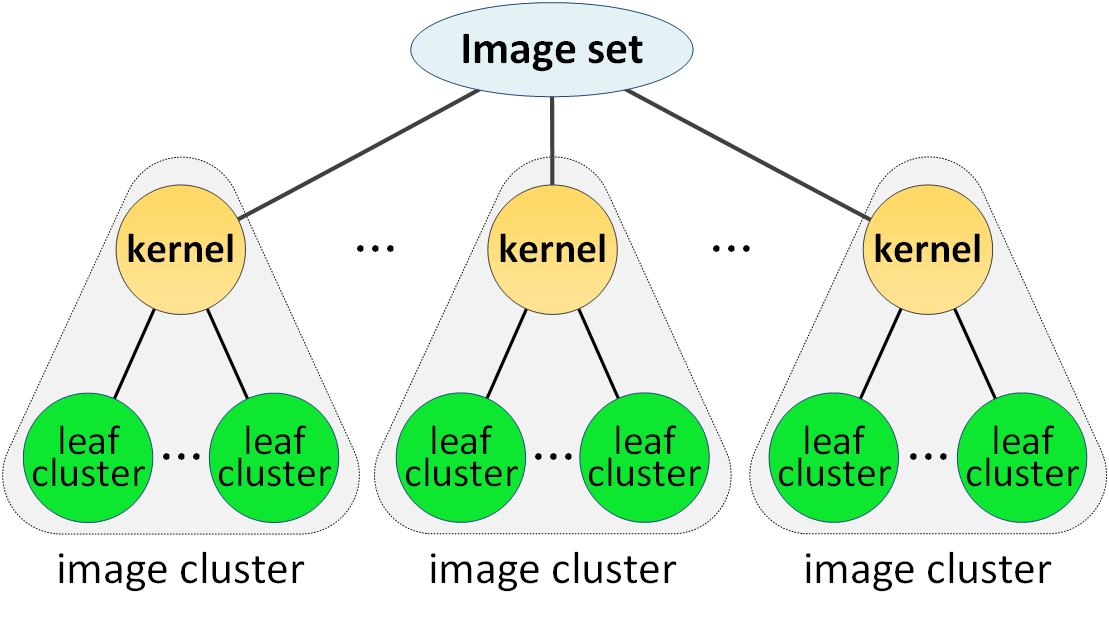}
	\end{center}
	\caption{The Trilaminar Multiway Reconstruction tree (TMR-tree).}
	\label{fig:tree}
\end{figure}

In this paper, we propose a new method for automatic data partitioning and multiple proper starting points selecting. The partitioning result is described with a Trilaminar Multiway Reconstruction tree (TMR-tree), which is shown in Fig. \ref{fig:tree}. A partition suitable for reconstruction must satisfy two requirements. On the one hand, each partition should contain a set of images having large overlap between each other. The reconstruction will start from these images to ensure accuracy. On the other hand, any two images in the same partition should be connected by scene overlap, so that all of them could be added. To this end, our method partitions the image set in two steps. We first search for places where images are densely distributed. Several kernels are found at these places, shown as yellow nodes in Fig. \ref{fig:tree}. Each kernel contains a few images having large overlap with each other. The number and size of kernels should not be too large. Images outside the kernels are called leaves. Next, all the images are clustered according to their optimal reconstruction paths to the kernels. Each cluster consists of two parts: one kernel and a set of leaves around it. Accordingly, the reconstruction performs in a hierarchical way. In the first stage, all the kernels are reconstructed in parallel to build base models of the scene. In the second stage, the leaves are added to these base models. Since kernels only take a small part of the whole image set, the number of leaves in a cluster might still be large. To enable faster speed, these leaves are further split into leaf clusters, which are green nodes in Fig. \ref{fig:tree}. Each leaf cluster can be independently added to the same base model in parallel without distinct accuracy deterioration. The models of leaf clusters sharing the same base model are merged to get the model of an image cluster. Then models of different image clusters are merged to a complete one.

\section{Related Works}\label{sec:related works}

\begin{figure*}[t]
	\begin{center}
		\includegraphics[width=0.83\linewidth]{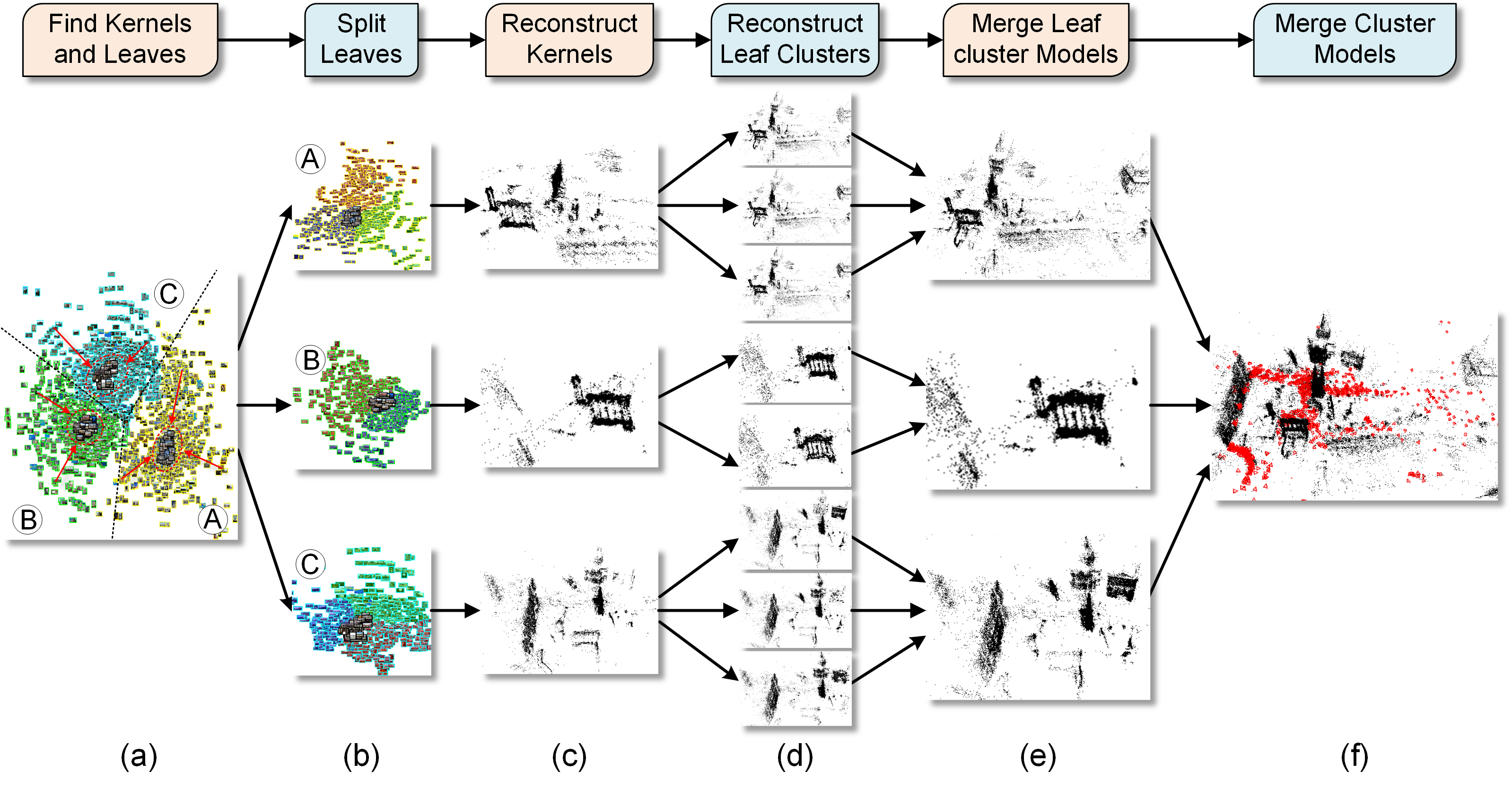}
	\end{center}
	\caption{The overall flowchart of our method. (a) The whole image set is divided into three clusters A, B and C. In each cluster the kernel is drawn with black boxes. The leaves in each image cluster are drawn with yellow, green and cyan boxes, respectively. (b) Leaves in an image cluster are further split into leaf clusters. (c) The base models reconstructed from the kernels. (d) Reconstruction results after adding each leaf cluster to the same base model. (e) The model of each image cluster after merging the models in (d). (f) The final result after merging the models of different image clusters.}
	\label{fig:mainidea}
\end{figure*}

Large scale structure from motion has witnessed great development in recent years.
Several complete structure from motion systems have been proposed. Snavely \textit{et al.} \cite{Snavely06,Snavely081} are among the first to propose a complete incremental SfM pipeline. The backbone of their \textit{Photo Tourism} system is a structure from motion approach which computes the photographers' locations and orientations, along with a sparse 3D point cloud. 

Li \textit{et al.} \cite{Xiaowei08} proposed to capture the major aspects of the scene using an iconic scene graph. Their method divided images into small clusters. Image matching and geometry verification are only performed between images within the same cluster. Each cluster is represented by an iconic image. In order to make SfM perform more efficiently, they partitioned the iconic scene graph with normalized cuts \cite{Shi00} and run incremental SfM on each part. 

Agarwal \textit{et al.} \cite{Agarwal09} designed a system running on a collection of parallel distributed machines to efficiently reconstruct a city. They paid a lot of effort to reduce the cost of scheduling between different tasks. They computed a skeletal set of photographs \cite{Snavely082} instead of reconstructing all the images. 
Frahm \textit{et al.} \cite{Frahm10} improved the work of \cite{Agarwal09} by reconstructing a city on a single machine with multi-core CPUs and GPUs. They concatenated the global GIST descriptor \cite{Oliva01} with a subsampled image. Then the descriptor was compressed to shorter binary code so that it is memory efficiency for GPU computation. They also generated dense 3D model using fast plane sweeping stereo and efficient depth map fusion algorithms. 

Wu \cite{Wu13} proposed a new SfM framework that has $O(n)$ time complexity. He used top-scale feature matching to coarsely identify image overlapping, which saved much time in image matching. During reconstruction, his method performed full bundle adjustment optimization after the model increases a certain ratio and partial bundle adjustment on a constant number of recently added cameras to reduce the accumulated time of bundle adjustment. 

Shah \textit{et al.} \cite{Shah15-2} proposed a coarse-to-fine SfM strategy. In the first stage a coarse yet global model is quickly reconstructed using high scale SIFT features. This model offers useful geometric constraints for the second stage, in which the model is enriched by localizing remaining images and triangulating remaining features. 

Heinly \textit{et al.} \cite{Heinly15} advanced the state-of-the-art SfM methods from city-scale modeling to world-scale modeling on a single computer. They also leverage the idea of iconic images to represent small image clusters. The database-side feature augmentation is applied so that an iconic image can cover a broader set of views. For the ability to handle world scale images, their system stores an image's data in memory only when it is needed. 

The latest achievement in large scale SfM is reported in \cite{schoenberger2016sfm}. Their work improved several components of the state-of-the-art methods, such as geometry verification, view selection, triangulation and bundle adjustment to make a further step towards a robust, accurate, complete and scalable system.

\section{Overview}\label{sec:overview}
In this section, an overview of the proposed method is given. Before our algorithm starts, some preparations such as feature extraction, image matching and matching graph construction are made. Suppose we have a set of unordered images $\textit{\textbf{I}}=\{I_i\}_{i=1}^{N}$. The SIFT \cite{Lowe04} features are extracted from each image. Each image is matched to its top $K$ nearest neighbors searched from a trained vocabulary tree. The value of $K$ is set to $30$ according to other reported papers. A faster GPU implementation \cite{siftgpu} is adopted to speedup the burdensome matching procedure. Wrong matches are removed by estimating the epipolar geometry between two views using the RANSAC \cite{Fischler1981} algorithm. 

After image matching is done, the matching graph is constructed. The matching graph $\textit{\textbf{G}}<\textit{\textbf{V}},\textit{\textbf{E}}>$ is an undirected weighted graph with a set of vertexes $\textit{\textbf{V}}$ and edges \textit{\textbf{E}}. A vertex $v_i \in \textit{\textbf{V}}$ represents an image. If two images have scene overlap, an edge is added between the corresponding vertexes. We build two kinds of matching graphs: a similarity graph $\textit{\textbf{S}}$ and a difference graph $\textit{\textbf{D}}$. They have the same number of vertexes and edges, but the meaning of their edge weights are different. In the similarity graph $\textit{\textbf{S}}$, the edge weight $s_{ij}$ reflects the content similarity between two images. An intuitive way is to measure this similarity with the number of matches between two images. However, this measurement is sensitive to image resolution and texture. High resolution or textured images will have more matches than low resolution or less textured images. In this paper, $s_{ij}$ is computed from the following formulation:
\begin{equation}\label{eq:S}
s_{ij}=\frac{n_{ij}}{n_i \cup n_j},
\end{equation}
in which $n_{ij}$ is the number of matches between two images $I_i$ and $I_j$, $n_i$ and $n_j$ are the number of feature points on image $I_i$ and $I_j$ that have corresponding points on the other images, respectively. Eq. \eqref{eq:S} is also known as the Jaccard similarity coefficient. A larger $s_{ij}$ indicates that $I_i$ and $I_j$ have more scene overlap. It is robust to different image sizes and scene textures. The weight of the difference graph $\textit{\textbf{D}}$ is then computed from:
\begin{equation}\label{eq:D}
d_{ij}=1-s_{ij}.
\end{equation}

The flowchart of the proposed method is shown in Fig. \ref{fig:mainidea}. In Fig. \ref{fig:mainidea}(a) three kernels are found and all the images are divided into three clusters A, B and C according to their optimal reconstruction paths to the kernels. Each cluster contains a kernel and some leaves around it. The kernels are drawn with black boxes and the leaves in different clusters are in yellow, green and cyan, respectively. Fig. \ref{fig:mainidea}(b) shows that the leaves in an image cluster are split into several leaf clusters to acquire faster reconstruction speed. The reconstruction path from each image in a leaf cluster to the kernel should lie within the same leaf cluster. Thus, each leaf cluster could be independently added to the same base model in parallel without distinct accuracy deterioration. After finishing the above steps, the kernels are first reconstructed in parallel to get several base models of the scene, which are shown in Fig. \ref{fig:mainidea}(c). These base models are reliable because they are reconstructed from images with large overlap. Next, different leaf clusters in one image cluster are independently added to the same base model in parallel, and the results are shown in Fig. \ref{fig:mainidea}(d). The base models are enriched by adding images in the leaf clusters in this step. Since these models share the same base model, it's easy to merge them together to get the model of each image cluster. The models for cluster A, B and C are shown in Fig. \ref{fig:mainidea}(e). Finally, the models of different image clusters are merged to a complete one, which is shown in Fig. \ref{fig:mainidea}(f).

\section{Trilaminar Multiway Reconstruction Tree}\label{sec:tree}
The data partitioning result is modeled by a Trilaminar Multiway Reconstruction Tree (TMR-tree). The top layer is a single root node representing the whole image set. The nodes in the middle and bottom layers correspond to kernels and leaf clusters. In this section, the method for building the Trilaminar Multiway Reconstruction Tree is introduced. The steps include: finding kernels, image clustering and finding leaf clusters.
\subsection{Finding Kernels with A Multi-layer Greedy Strategy}\label{subsec:kernel}
Kernels are used to reconstruct base models of the scene. They should be found at places where images are densely distributed. Images at such places have large scene overlap between each other, so that the base model reconstructed is accurate in precision, representative and centric in location. Since there is no absolute standard for judging whether the distribution of the cameras is dense, we choose a loose greedy manner to progressively find multiple kernels from the whole image set.

Since the mission of a kernel is to reconstruct an initial local model of the scene, it is not expected to contain too many images. Suppose the ideal size of a kernel is between $m$ and $\alpha\cdotp m$, where $m$ is a positive number and $\alpha\geq1$ is an inflation factor. We adopt a greedy strategy to find kernels in a layered graph. Given the number of layers $k$ and a set of edge wight thresholds $\theta_i(i\in {1,2,\dots,k})$ satisfying $\theta_i>\theta_{i+1}$, the edges whose weights are greater than $\theta_i$ are added back to the similarity graph $\textit{\textbf{S}}$ in the $i^{th}$ step. Then we find connected components in the graph. If none of them is larger than $m$, we continue by reducing the edge weight threshold and adding more edges in the next step. If a connected component is larger than $m$ but smaller than $\alpha\cdotp m$, the images in this component form a new kernel and the corresponding vertexes are removed from the current graph. If a connected component is larger than $\alpha\cdotp m$, we will find kernels with proper size in this component recursively using the same method above. In this way, we can guarantee that each kernel is a set of images that have the strongest overlap among the current remaining images.

\begin{figure}[t]
	\begin{center}
		\includegraphics[width=0.55\linewidth]{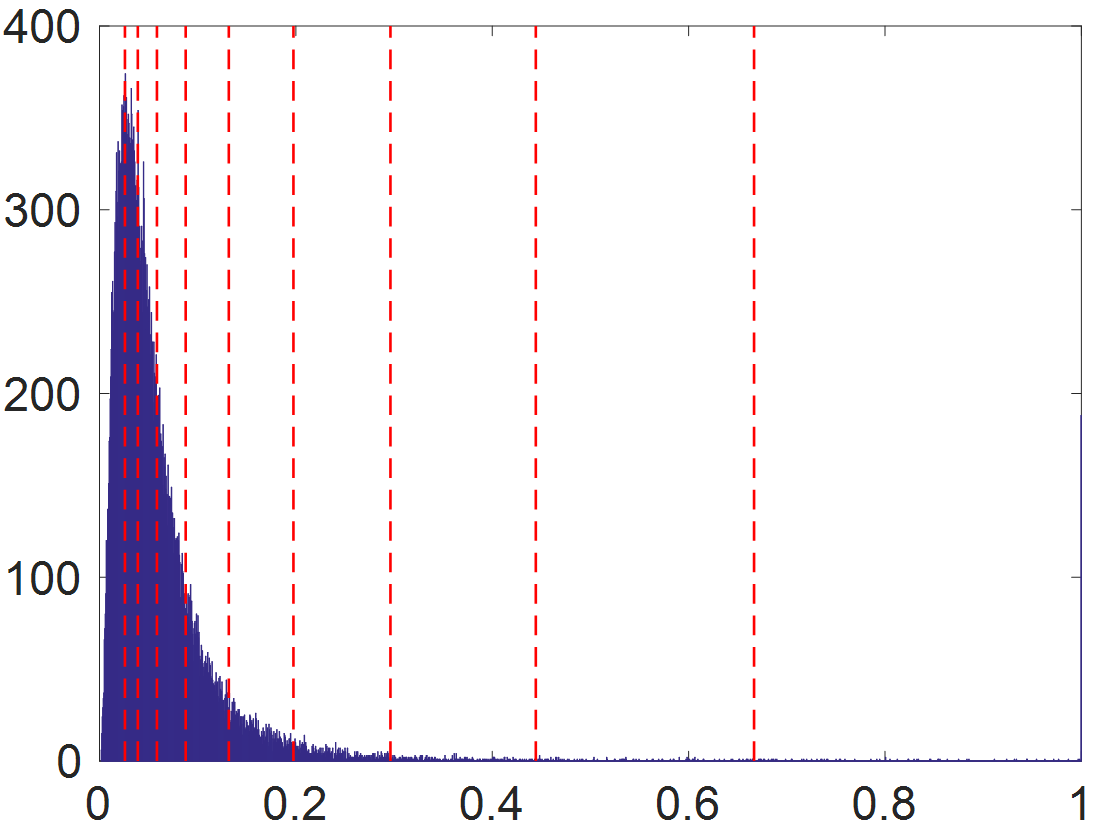}
	\end{center}
	\caption{The distribution of edge weights in the similarity graph and $\theta_i$.}
	\label{fig:edge_weight}
\end{figure}

Computing $\theta_i$ is an important problem. Denote the minimum and maximum edge weights in the similarity graph as $a$ and $b$, respectively. In practice, $a$ is set to a value $\varepsilon$ larger than the minimum of the edge weights so that images having too weak overlap with others are not considered in this stage. The range $[a,b]$ is divided into $k$ intervals and edges are added from higher interval to lower interval. Fig. \ref{fig:edge_weight} shows the distribution of all the edge weights in the similarity graph. It can be seen that there is an obvious peak near 0.02. If the intervals are divided uniformly, the higher intervals contain few edges but the lower intervals contain numerous edges. As a result, it is difficult to find large enough kernels at the first few steps but will soon fall into deep recursion because adding a great many edges will make the connected component grow fast. In this paper, $\theta_i$ is computed from the following formulation:
\begin{equation}\label{eq:interval}
\theta_i=a+\frac{b-a}{1.5^{i-1}}, i\in {1,2,\dots,k},
\end{equation}
which are red vertical dash lines in Fig. \ref{fig:edge_weight}.
It can be seen that such a division can keep the number of edges in each interval roughly the same.

Once several kernels have been found, an exemplar image which will be used as the starting point is found in each kernel. It should have dense overlap with other images so that the initial model can easily spread the 3D structure to nearby space. The Affinity Propagation (AP) clustering algorithm \cite{Frey07} is applied to images in each kernel. All the centers and their adjacent neighbors on the similarity graph $\textit{\textbf{S}}$ are treated as the candidates for the exemplar image. Affinities between data points required by AP clustering are computed from Eq. \eqref{eq:S}. The reason for choosing AP clustering has two aspects. On the one hand, AP clustering algorithm can automatically determine the number of clusters. On the other hand, the center of a cluster is one data point instead of a virtual mean position. For each candidate image, the following score is computed:
\begin{equation}\label{eq:score_startingpoint}
\delta(v)=h_{deg}(v)+\beta_1\cdot h_{sim}(v)+\beta_2\cdot h_{ndeg}(v).
\end{equation}
The first term $h_{deg}(v)$ is the degree of the vertex $v$, which counts the number of images that overlap with it. The second term $h_{sim}(v)$ is the average similarity from the vertex $v$ to its neighbors, namely the mean adjacent edge weight on $\textit{\textbf{S}}$. This term encourages the vertex $v$ to have large overlap with its neighbors. The last term $h_{ndeg}(v)$ is the average degree for the neighbors of $v$. That is to say, not only $v$ itself should overlap with many images, but also the images overlapping with it should also overlap with as many other images as possible. This strengthens the potential of the starting point to build an accurate initial model and spread 3D structure to nearby space. Image with the highest score is selected as the exemplar image. 

\subsection{Clustering Images According to Their Optimal Reconstruction Paths to the Kernels}\label{subsec:subordination}
\begin{figure}[t]
	\begin{center}
		\includegraphics[width=0.85\linewidth]{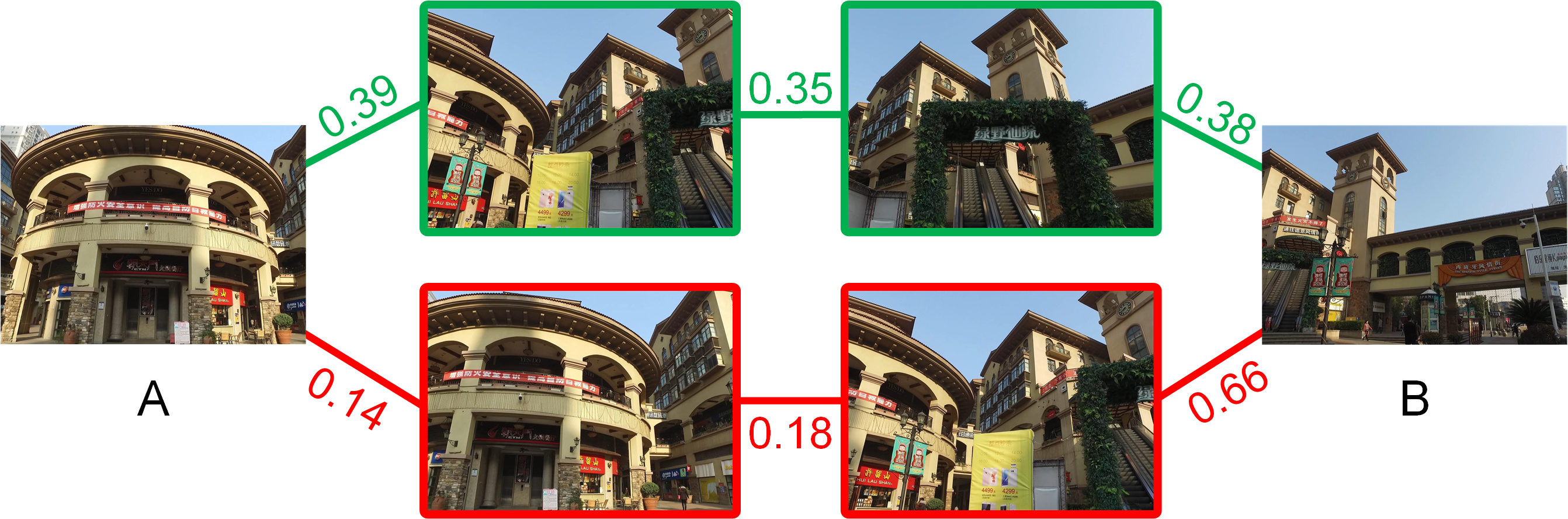}
	\end{center}
	\caption{Find the optimal path (in green) using the Multi-layer Shortest Path (MSP) algorithm.}
	\label{fig:multi_layer_shortest_path}
\end{figure}

In this part, all the images are clustered by treating the kernels as centers. This is not a simple image classification problem because the clusters may not be good for reconstruction. Hence, in our method all the images are clustered according to their optimal reconstruction paths to the kernels. A reconstruction path is composed of a series of overlapping images which can pass the 3D structure. There can be multiple reconstruction paths from an image to a kernel. We think the optimal reconstruction path should consist of a series of largely and equally overlapping images. In other words, the maximum difference between adjacent images on an optimal reconstruction path should be minimized, which is shown in Fig. \ref{fig:multi_layer_shortest_path}. In this example, two reconstruction paths between images A and B are shown. The edge weights reflects the difference between two images. The red path has shorter length than the green path. However, it is not considered as the optimal reconstruction path because its edge weights vary a lot. There is an edge whose weight (0.66) is much larger than the other two edges (0.14 and 0.18). This means that the 3D structure has to be propagated via relatively weak image overlap, which is unreliable. Although the length of the green path is a bit longer than the red path, its edge weights are similar. If the green path is selected as the optimal reconstruction path, the risk of passing 3D structure via weak overlap will no longer exist.

In this paper, a Multi-layer Shortest Path (MSP) algorithm is proposed to find the optimal reconstruction paths from each image to the kernels. Our MSP algorithm operates on the difference graph $\textit{\textbf{D}}$, in which the edge weight indicates the scene difference between images. This graph is divided into $L$ layers by a set of increasing weight thresholds $\phi_t(t \in {1,\dots,L})$ satisfying $\phi_t<\phi_{t+1}$. More specifically, the range of edge weights $[\min(d_{ij}),\max(d_{ij})]$ on $\textit{\textbf{D}}$ is divided into $L$ homogeneous intervals. For each interval the step length is $l=(\max(d_{ij})-\min(d_{ij}))/L$ and $\phi_t$ is computed from
\begin{equation}\label{eq:layer_MSP}
\phi_t=t*l+\min(d_{ij}),t=1,\dots,L.
\end{equation}
Edges whose weights are smaller than $\phi_t$ are added back to $\textit{\textbf{D}}$ in the $t^{th}$ layer. Denote $w$ as a leaf. The shortest paths from $w$ to the exemplars of the kernels are computed. At the very beginning, no paths exist between $w$ and the kernels. If none of the paths to the kernels are found in the $t^{th}$ layer, we then add more edges by using a larger edge weight threshold in the next layer. With more and more edges are added, the paths between $w$ and the kernels will be found. The optimal path between $w$ and a kernel is the path found for the first time. Although in the following layers the paths between $w$ and the same kernel will also be found, they are not optimal. In the example of Fig. \ref{fig:multi_layer_shortest_path}, the optimal path in green will be found before the path in red. If the optimal paths to different kernels are found in the $t^{th}$ layer, then $w$ is assigned to the cluster of the kernel with the smallest path length. Once a leaf has been clustered, it will not be processed in the following layers. 

\subsection{Finding Leaf Clusters using Radial Agglomerate Clustering}\label{subsec:splitting}
Since kernels take only a small part of the image set, the number of leaves in an image cluster might be still too large. Adding them sequentially to the base model will be time consuming. Thus, they are further divided into leaf clusters to achieve faster speed. Three conditions should be satisfied so that each leaf cluster could be reconstructed in parallel without distinct accuracy deterioration. (1) Images within each leaf cluster should have considerable overlap with each other. (2) Each leaf cluster should have strong overlap with the kernel so that it can be added to the base model. (3) The size for these leaf clusters should be balanced to reduce waiting time of different threads. 

\begin{figure}[t]
	\begin{center}
		\includegraphics[width=\linewidth]{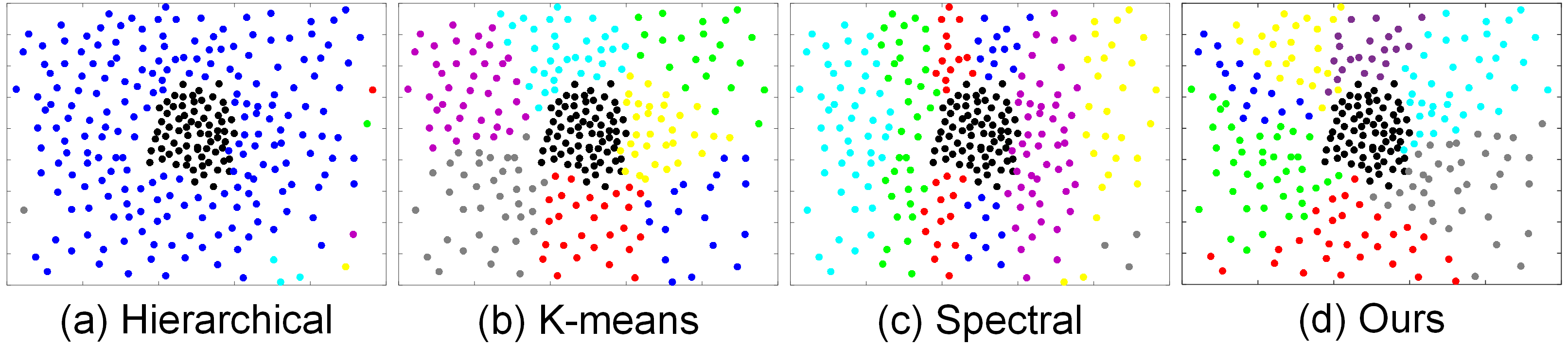}
	\end{center}
	\caption{The results of Hierarchical clustering, K-means clustering, Spectral clustering and the proposed Radial Agglomerate Clustering (RAC) algorithm on 2D synthetic data. The black points in the center are manually selected kernel points.}
	\label{fig:Radial_Agglomerate_Clustering}
\end{figure}

In this paper, an improved Radial Agglomerate Clustering (RAC) algorithm is proposed to divide leaf clusters. The distance between leaves in an image cluster is the shortest path length on the subgraph formed by this image cluster. The distance between two leaf clusters is measured by the distance of two closest images between them. The number of leaf cluster $K_c$ is computed from $K_c=round(\frac{M}{e})$, where $M$ is the number of leaves in an image cluster and $e$ is the ideal mean size of each leaf cluster. We expect a leaf cluster to be larger than the kernel and set $e=r\cdot m$, where $r$ is a positive integer. At the very beginning, each leaf in the image cluster is an isolate leaf cluster. In each step a score is computed for ever possible leaf cluster pair and two leaf clusters with the smallest score are merged until $K_c$ leaf clusters remain. The score is computed from:
\begin{equation}\label{eq:score_split}
\varphi(p)=\sigma_1\cdotp g_{d}(p)+\sigma_2\cdotp g_{k}(p)-\sigma_3\cdotp g_{r}(p)+\sigma_4\cdotp g_{c}(p)，
\end{equation}
where $p$ is a possible leaf cluster pair composed of two leaf clusters $c_1$ and $c_2$. $\sigma_1$, $\sigma_2$, $\sigma_3$ and $\sigma_4$ are four positive tunning parameters. The first term $g_{d}(p)$ is the distance between $c_1$ and $c_2$, which prefers to merge close leaf clusters to meet requirement (1). The second term $g_{k}(p)$ measures the distance between the kernel and the leaf cluster after merging $c_1$ and $c_2$. This term encourages that after merging two leaf clusters the new one has strong connection with the kernel. The third term $g_{r}(p)$ is the difference between the distances from the two leaf clusters to the kernel. It will guide the merging along the radial direction. The second and third terms act together to meet requirement (2). The last term $g_{c}(p)$ counts the cardinality of $c_1$ and $c_2$ after merging them to a new one. It will tend to merge small leaf clusters at each step so that our final leaf clusters are balanced in size, which satisfies requirement (3). 

An example showing the clustering results of our RAC algorithm and several other methods such as Hierarchical clustering, K-means clustering and Spectral clustering on synthesized 2D points is in Fig. \ref{fig:Radial_Agglomerate_Clustering}. Kernel points are in the center with black color. The remaining points are divided into 7 clusters by different methods. It can be seen that the result of hierarchical clustering is unbalanced. Some small clusters are far from the kernel points. The K-means clustering algorithm produces nearly balanced clusters. But the green and blue clusters are not adjacent with the kernel points. Similar problem happens to the spectral clustering algorithm as well. Our RAC method can produce radial, compact and balanced clusters.

\section{Parallel Reconstruction with the TMR-tree}\label{sec:reconstruction}
The reconstruction performs in two stages. In the first stage, the kernels are reconstructed in parallel to get several base models. A kernel is reconstructed from two initial images. The first image is fixed to the exemplar image found in Sec. \ref{subsec:kernel}. The second image is set to the one having the most matches and relatively wide baseline with the first image. In the second stage, leaf clusters of a kernel are added to the same base model, producing several individual models in parallel. The pose of each image is initialized by solving the PnP problem and then refined via bundle adjustment. At last, the individual models are merged in two steps. First, the models of different leaf clusters sharing the same kernel are merged to get the model of an image cluster. Next, different image cluster models returned in the first step are merged to get a complete model of the scene.

The reconstruction can be very fast if we have enough CPU cores and GPU cards because all the kernels and leaf clusters can be reconstructed in parallel. The complexity of our method is relevant to the kernel size $\alpha\cdot m$ and the leaf cluster size $r\cdot m$, which is $O(m)$. While even a linear-time SfM algorithm \cite{Wu13} has a complexity of $O(N)$, where $N$ is the number of cameras. Our method has a theoretical speedup factor of $\frac{N}{m}$. If $m$ increases, the complexity will increase but the models are more stable. On very large image set, the difference between $N$ and $m$ is large and the speedup is more obvious.

A similarity transformation is computed to merge two models. One of the difficulties is to detect the common parts between them. Since the same track reconstructed in different models may be inconsistent, directly finding common 3D points between models according to shared tracks will include a very large portion of outliers. In this paper we narrow down the number of suspicious common 3D points by the following method. Consider two models $M_1$ and $M_2$, our method first finds an image in $M_2$ who has the most tracks reconstructed in $M_1$. Then the tracks on this image who have also been reconstructed in $M_2$ are counted. If the number of such tracks is greater than a threshold $\tau$, a Least-Square method \cite{Umeyama91} is implemented in the RANSAC framework to robustly estimate the similarity transformation between them. Otherwise do not merge $M_1$ and $M_2$. 

\section{Experiment Results}\label{sec:experiments}
\subsection{Parameter Settings and Implementation details}\label{subsec:parameters}
Among all the parameters, the most important two are $m$ and $\varepsilon$ because they affect the granularity of our data partition. A larger $m$ will reduce the speed of the algorithm but the result might be more stable. A smaller $m$ will result in fragmented partitions and inaccurate models. Similarly, increasing $\varepsilon$ will reduce the number of kernels and reducing $\varepsilon$ will result in more kernels. In this paper, we set $m=min(70, 0.15*Z)$, where $Z$ is the total number of images, and $\varepsilon=0.1$ empirically. For the rest parameters, we set $\alpha=1.5$, $k=15$, $\beta_1=100$, $\beta_2=1$, $r=3$, $\sigma_1=1$, $\sigma_2=1$, $\sigma_3=3$, $\sigma_4=1$ and $\tau=12$.

The GPU based bundle adjustment algorithm \cite{Changchang_pba} is used. Our algorithm is implemented using C++ on Ubuntu 14.10 operating system. The experiments are tested on a machine with two Intel Xeon CPU E5-2630 v3 2.40GHz, one NVIDIA GeForce GTX TitanX graphics card and 256GB RAM.

\begin{figure}[t]
	\begin{center}
		\includegraphics[width=0.8\linewidth]{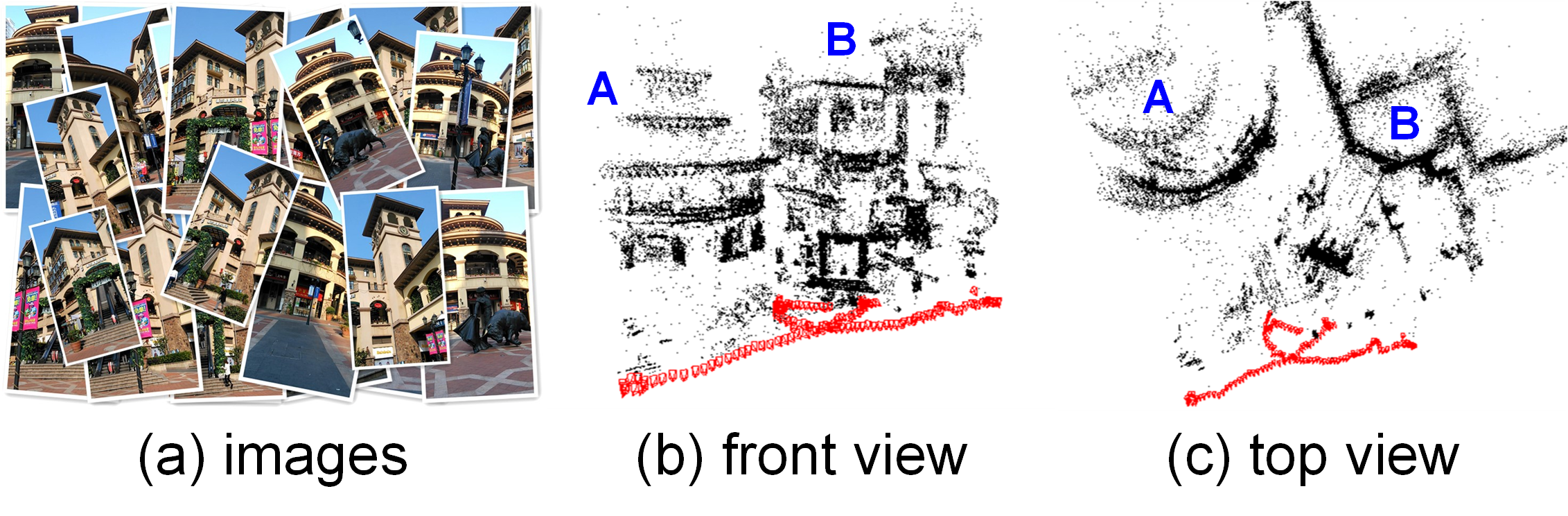}
	\end{center}
	\caption{(a) The images captured in a street scene with a hand held digital camera. (b) Front view of its sparse 3D model. (c) Top view of its sparse 3D model.}
	\label{fig:toy1}
\end{figure}

\begin{figure}[t]
	\begin{center}
		\includegraphics[width=0.88\linewidth]{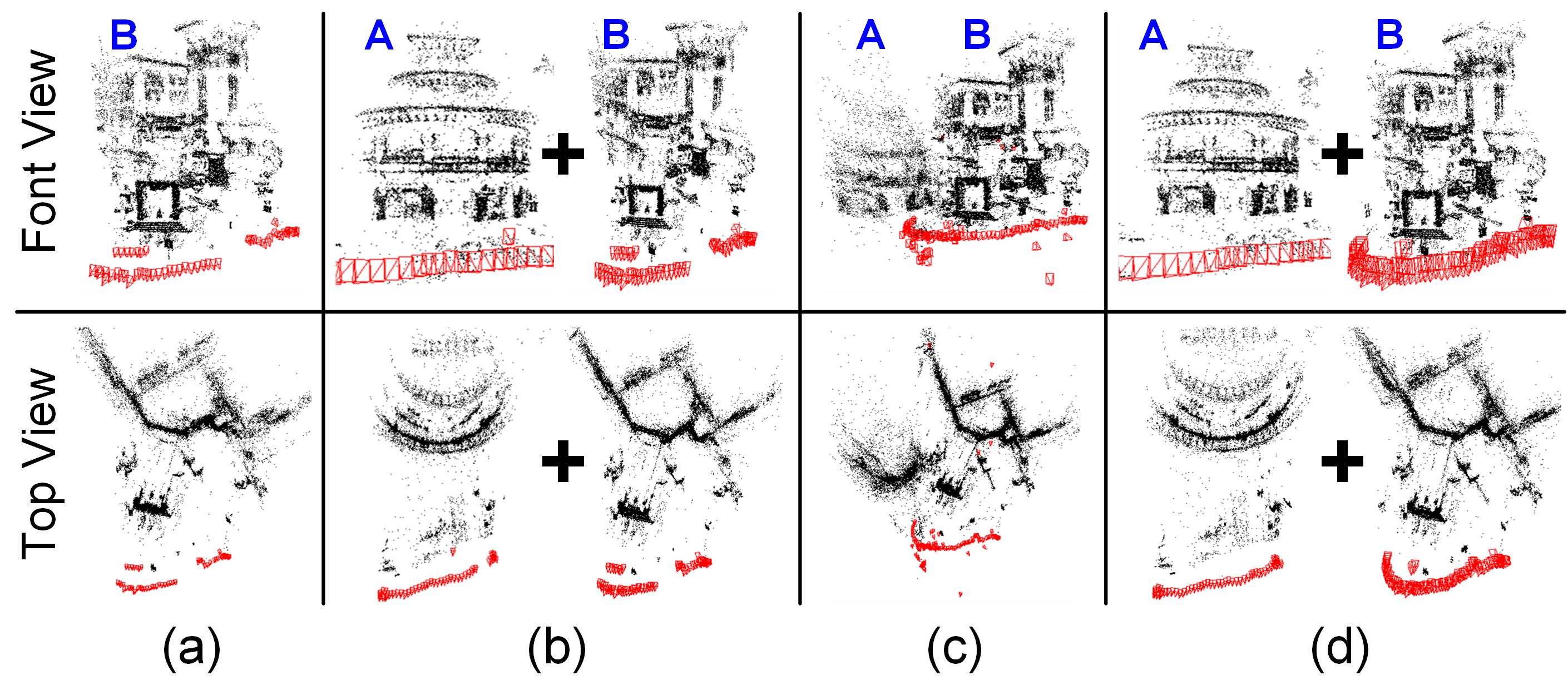}
	\end{center}
	\caption{The reconstruction results of Bundler \cite{Snavely081} and our method on self-captured images. (a) and (b) are results of Bundler and our method after removing 35 images. (c) and (d) are results of Bundler and our method after adding 21 images back.}
	\label{fig:toy2}
\end{figure}

\subsection{Results on Self-captured Images}\label{subsec:self-capture}
In this part we show the results on self-captured images. This image set contains 135 images of a scene beside the street. It contains two buildings A and B. When taking these images, we require adjacent pictures to have sufficient overlap so that all images could be used to reconstruct a complete model. Fig. \ref{fig:toy1} (a) shows some example images. The sparse 3D point cloud together with the poses of all the cameras are shown in Fig. \ref{fig:toy1} (b) and (c).

We first remove 35 images between A and B to cut the whole image set into two isolated parts. The reconstruction results for Bundler and our method are shown in Fig. \ref{fig:toy2}(a) and (b), respectively. Our method can reconstruct two independent models while Bundler reconstructs only one of them. Then, 21 images are added back to the image set so that there is weak overlap between A and B. The result of Bundler in Fig. \ref{fig:toy2}(c) shows that 3D structure is passed from B to A, and all the cameras are reconstructed. However, the structure and camera poses are wrong because the overlap between A and B are unreliable. In this case, it's better to build several good partial models from image subsets rather than build a wrong model with all the images. As is shown in Fig. \ref{fig:toy2}(d), our method divides the image set into two clusters and builds two correct independent models for A and B. If more images in the middle are provided, the two independent models will be merged to a complete one in the right way.

\subsection{Results on Public Benchmarks}\label{subsec:public-benchmarks}

\begin{table*}[t]
	\centering
	\caption{Partition result on the Montreal Notre Dame, Vienna Cathedral and Yorkminster image sets. For each dataset, the number of kernels, the number of leaf clusters belonging to a kernel and the time are given.} \label{tab:public_dataset2_partition}
	\small
	\begin{tabular}{|c|c|c|c|c|c|c|c|c|c|c|c|c|c|}
		\hline
		\textbf{Dataset} & \multicolumn{4}{c|}{Montreal Notre Dame} & \multicolumn{5}{c|}{Vienna Cathedral} & \multicolumn{4}{c|}{Yorkminster} \\ \hline
		
		\textbf{Kernels} & K 1 & K 2 & K 3 & K 4 & K 1 & K 2 & K 3 & K 4 & K 5 & K 1 & K 2 & K 3 & K 4 \\ \hline
		
		\textbf{Num Leaf Clusters} & 3 & 2 & 1 & 1 & 3 & 1 & 1 & 1 & 2 & 1 & 1 & 1 & 1 \\ \hline
		
		\textbf{Time} & \multicolumn{4}{c|}{7.127s} & \multicolumn{5}{c|}{33.107s} & \multicolumn{4}{c|}{48.324s} \\ \hline
	\end{tabular}
\end{table*}

\begin{table*}[t]
	\centering
	\caption{Results on the Montreal Notre Dame, Vienna Cathedral and Yorkminster datasets. For each model, the number of reconstructed cameras and the mean reprojection error are given. The running time for reconstruction is in the last column.} \label{tab:public_dataset2_result}
	\small
	\begin{tabular}{c|c|c|c|c|c|c|c|c}
		\hline
		\textbf{Dataset} & \textbf{Method} & \multicolumn{3}{c|}{\textbf{\#Cameras}} & \multicolumn{3}{c|}{\textbf{Error (pixel)}} & \textbf{Time} \\ \hline
		
		\multirow{5}{*}{Montreal Notre Dame} & \multirow{2}{*}{Ours} & Model 1 & Model 2 & Model 3 & Model 1 & Model 2 & Model 3 & \multirow{2}{*}{217.2s} \\ \cline{3-8}
		& & 385 & 355 & 97 & 0.6241 & 0.7286 & 0.5112 & \\ \cline{2-9}
		
		& \multirow{2}{*}{VisualSFM} & Model 1 & Model 2 & Model 3 & Model 1 & Model 2 & Model 3 & \multirow{2}{*}{457s} \\ \cline{3-8}
		& & 343 & 504 & 97 & 1.596 & 1.467 & 0.909 & \\ \cline{2-9}
		
		& Bundler & - & 399 & - & - & 1.5083 & - & 648.2s
		\\ \hline\hline
		
		\multirow{5}{*}{Vienna Cathedral} & \multirow{2}{*}{Ours} & \multicolumn{2}{c|}{Model 1} & Model 2 & \multicolumn{2}{c|}{Model 1} & Model 2 & \multirow{2}{*}{337.4s} \\ \cline{3-8}
		& & \multicolumn{2}{c|}{1000} & 292 & \multicolumn{2}{c|}{0.6550} & 0.8684 & \\ \cline{2-9}
		
		& \multirow{2}{*}{VisualSFM} & \multicolumn{2}{c|}{Model 1} & Model 2 & \multicolumn{2}{c|}{Model 1} & Model 2 & \multirow{2}{*}{1216s} \\ \cline{3-8}
		& & \multicolumn{2}{c|}{929} & 275 & \multicolumn{2}{c|}{1.901} & 1.519 & \\ \cline{2-9}
		
		& Bundler & \multicolumn{2}{c|}{1197} & - & \multicolumn{2}{c|}{0.7106} & - & 12181.2s
		\\ \hline\hline
		
		\multirow{5}{*}{Yorkminster} & \multirow{2}{*}{Ours} & Model 1 & Model 2 & Model 3 & Model 1 & Model 2 & Model 3 & \multirow{2}{*}{282.7s} \\ \cline{3-8}
		& & 593 & 333 & 121 & 0.6935 & 0.5451 & 0.5905 & \\ \cline{2-9}
		
		& \multirow{2}{*}{VisualSFM} & Model 1 & Model 2 & Model 3 & Model 1 & Model 2 & Model 3 & \multirow{2}{*}{796s} \\ \cline{3-8}
		& & 517 & 128 & 106 & 1.429 & 0.639 & 0.664 & \\ \cline{2-9}
		
		& Bundler & - & - & 122 & - & - & 0.6265 & 209.3s
		\\ \hline
	\end{tabular}
\end{table*}

\begin{figure}[t]
	\begin{center}
		\includegraphics[width=0.9\linewidth]{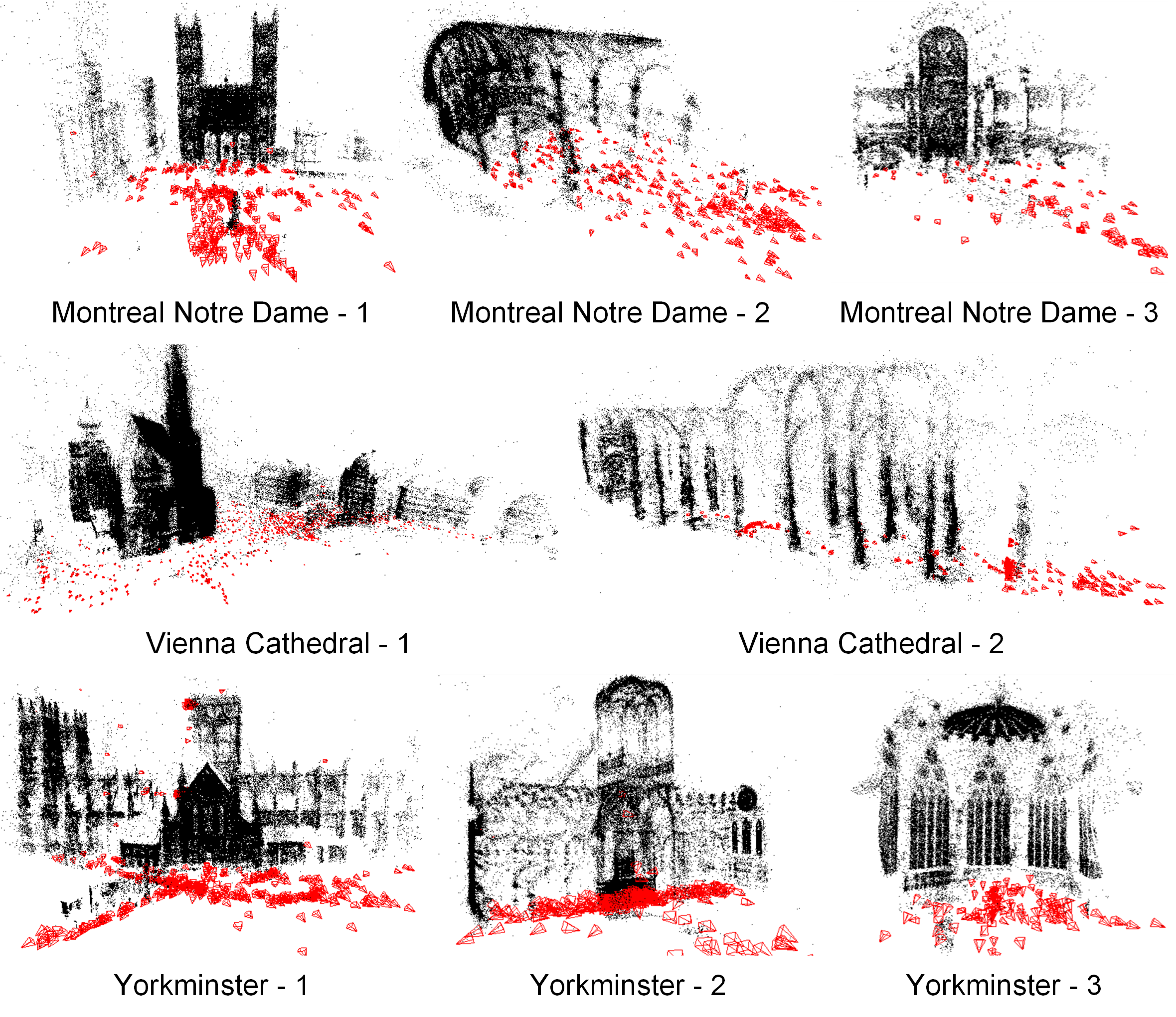}
	\end{center}
	\caption{The reconstruction results of our method. From top to bottom are: 3 models in Montreal Notre Dame, 2 models in Vienna Cathedral and 3 models in Yorkminster, respectively.}
	\label{fig:public_dataset2}
\end{figure}

Then the results on three public benchmarks including
Montreal Notre Dame \cite{Wilson14}, Vienna Cathedral \cite{Wilson14} and Yorkminster \cite{Wilson14} are reported. The number of images in these datasets are 2298, 6288 and 3368, respectively. Images in these image sets are not connected and contains several independent primary models. Fig. \ref{fig:public_dataset2} shows our reconstruction results on these datasets. We have found three primary models for Montreal Notre Dame, two primary models for Vienna Cathedral and three primary models for Yorkminster. Table \ref{tab:public_dataset2_partition} presents the partition results of our method on these image sets. The partition are done quickly within 1 minute. The number of kernels and leaf clusters for the three datasets are 4, 5, 4 and 10, 8, 4, respectively.

Our method is compared with two state-of-the-art methods: Bundler \cite{Snavely081} and VisualSFM \cite{Changchang_visualsfm}. The number of reconstructed cameras, the mean reprojection error and the running time for reconstruction are given in Table \ref{tab:public_dataset2_result}.  Since Bundler finds a single starting point and runs incremental SfM once, it can only reconstruction one of the models. The result of VisualSFM contains dozens of models because it iteratively runs a new incremental SfM in the remaining images after one model is reconstructed. However, most models are too small and only the primary models same with ours are considered here. The number of cameras reconstructed can reflect the model completeness of a SfM algorithm. For the total number of reconstructed cameras in each dataset, our method is the best. On a single model, the number of cameras reconstructed by us is similar with Bundler and larger than VisualSFM in most cases. The mean reprojection error indicates whether an algorithm is accurate. Our method achieves the smallest reprojection error on all the models.

The running time for reconstruction is in the last column of Table \ref{tab:public_dataset2_result}. It contains two parts: time for solving the PnP problem and time for bundle adjustment. For the first two datasets, Bundler takes the longest time because bundle adjustment is not performed on GPU. It runs the fastest on the third dataset because only a very small model is reconstructed. Both VisualSFM and our method can reduce much time by using the GPU based bundle adjustment, while the PnP solver is still implemented on CPU. However, VisualSFM runs the PnP solver serially while our algorithm runs it on different kernels or leaf clusters in parallel. So our algorithm is 2-4 times faster than VisualSFM. Theoretically the GPU bundle adjustment can be parallelized on different kernels or leaf clusters as well. But our machine has only one GPU card. Hence, bundle adjustment is actually executed serially in our method. When dealing with very large image set containing hundreds of kernels and leaf clusters, if we have enough CPU cores and multiple GPU cards, all the kernels and leaf clusters could be truly reconstructed in parallel and the speedup will be more remarkable.

\section{Conclusion}\label{sec:conclustion}
In this paper, an image set partitioning and starting point selecting method is proposed for efficient large scale SfM. The whole image set is divided into several clusters. Each image cluster consists of a kernel and a set of leaf clusters. A Trilaminar Multiway Reconstruction Tree (TMR-tree) is proposed to represent the partition result. The kernels are reconstructed first in parallel to build base models of the scene, and different leaf clusters of a kernel are added to the same base model simultaneously for parallel reconstruction. Experiments show that our method achieves much faster speed, more accurate poses and more complete models than state-of-the-art methods.  

{\small
	\bibliographystyle{ieee}
	\bibliography{sunbird}
}

\end{document}